\documentclass[runningheads]{llncs}
\usepackage{graphicx}
\usepackage{tikz}
\usepackage{comment} 
\usepackage{amsmath,amssymb} % define this before the line numbering.
\usepackage{color}
\usepackage{subfigure}
\usepackage{tabularx}
\usepackage{multirow}
\usepackage[rightcaption]{sidecap}
\usepackage{wrapfig}

\newcommand{\figref}[1]{Fig. \ref{#1}}

\makeatletter
\def\hlinewd#1{%
	\noalign{\ifnum0=`}\fi\hrule \@height #1 \futurelet
	\reserved@a\@xhline}

\begin{document}
% \linenumbers
\pagestyle{headings}
\mainmatter
\def\ECCVSubNumber{6316}  % Insert your submission number here

\title{Volumetric Transformer Networks} % Replace with your title

\begin{comment}
\titlerunning{ECCV-20 submission ID \ECCVSubNumber} 
\authorrunning{ECCV-20 submission ID \ECCVSubNumber} 
\author{Anonymous ECCV submission}
\institute{Paper ID \ECCVSubNumber}
\end{comment}
%******************

%\begin{comment}
\titlerunning{Volumetric Transformer Networks}
\author{Seungryong Kim\inst{1} \and
Sabine S\"{u}sstrunk\inst{2} \and
Mathieu Salzmann\inst{2}}
\authorrunning{S. Kim et al.}
\institute{Department of Computer Science and Engineering, Korea University, Korea\\
\email{seungryong\_kim@korea.ac.kr} \and
School of Computer and Communication Sciences, EPFL, Switzerland\\
\email{\{sabine.susstrunk, mathieu.salzmann\}@epfl.ch}}
%\end{comment}
%******************
\maketitle

\begin{abstract}
Existing techniques to encode spatial invariance within deep convolutional neural networks (CNNs) apply the same warping field to all the feature channels. This does not account for the fact that the individual feature channels can represent different semantic parts, which can undergo different spatial transformations w.r.t. a canonical configuration. To overcome this limitation, we introduce a learnable module, the volumetric transformer network (VTN), that predicts channel-wise warping fields so as to reconfigure intermediate CNN features spatially and channel-wisely. We design our VTN as an encoder-decoder network, with modules dedicated to letting the information flow across the feature channels, to account for the dependencies between the semantic parts. We further propose a loss function defined between the warped features of pairs of instances, which improves the localization ability of VTN. Our experiments show that VTN consistently boosts the features' representation power and consequently the networks' accuracy on fine-grained image recognition and instance-level image retrieval.
\end{abstract}

\section{Introduction}
Learning discriminative feature representations of semantic object parts is key to the success of computer vision tasks such as fine-grained image recognition~\cite{Fu17,Zheng17}, instance-level image retrieval~\cite{Noh17,Radenovic18}, and people re-identification~\cite{Zheng15,Li18}. This is mainly because, unlike generic image recognition and retrieval~\cite{imagenet_cvpr09,Everingham15}, solving these tasks requires handling subtle inter-class variations.

A popular approach to extracting object part information consists of exploiting an attention mechanism within a deep convolutional neural network (CNN)~\cite{Gregor15,Xu15,Noh17,Woo18}. While effective at localizing the discriminative parts, such an approach has limited ability to handle spatial variations due to, e.g., scale, pose and viewpoint changes, or part deformations, which frequently occur across different object instances~\cite{Felzenszwalb10,Jaderberg15,Dai17}. To overcome this, recent methods seek to spatially warp the feature maps of different images to a canonical configuration so as to remove these variations and thus facilitate the subsequent classifier’s task. This trend was initiated by the spatial transformer networks (STNs)~\cite{Jaderberg15}, of which many variants were proposed, using a recurrent formalism~\cite{Lin17inv}, polar transformations~\cite{Esteves18}, deformable convolutional kernels~\cite{Dai17}, and attention based samplers~\cite{Recasens18,Zheng19}. All of these methods apply the \emph{same} warping field to \emph{all} the feature channels. This, however, does not account for the findings of~\cite{Ren15,Bau17,Garcia18}, which have shown that the different feature channels of standard image classifiers typically relate to different semantic concepts, such as object parts. Because these semantic parts undergo different transformations w.r.t. the canonical configuration, e.g., the wings of a bird may move while its body remains static, the corresponding feature channels need to be transformed individually.

%As such, they cannot handle severe geometric deformations, where different semantic parts undergo different transformations.

\begin{figure}[t]
	\centering
	\renewcommand{\thesubfigure}{}
	\subfigure[]
	{\includegraphics[width=0.163\linewidth]{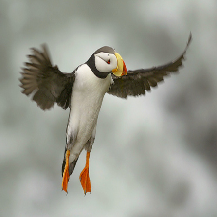}}\hfill	
	\subfigure[]
	{\includegraphics[width=0.163\linewidth]{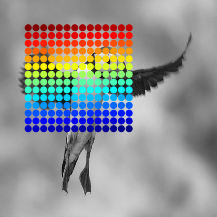}}\hfill	
	\subfigure[]
	{\includegraphics[width=0.163\linewidth]{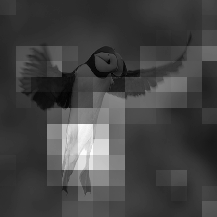}}\hfill	
	\subfigure[]
	{\includegraphics[width=0.163\linewidth]{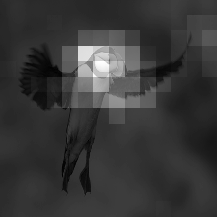}}\hfill	
	\subfigure[]
	{\includegraphics[width=0.163\linewidth]{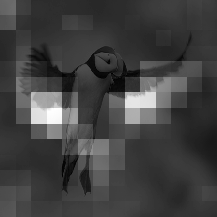}}\hfill	
	\subfigure[]
	{\includegraphics[width=0.163\linewidth]{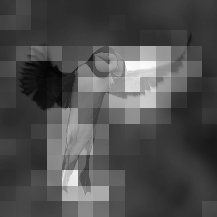}}\hfill\\	
	\vspace{-20.5pt}
	\subfigure[(a)]
	{\includegraphics[width=0.163\linewidth]{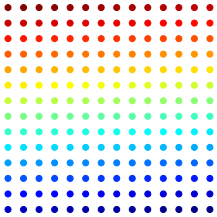}}\hfill	
	\subfigure[(b)]
	{\includegraphics[width=0.163\linewidth]{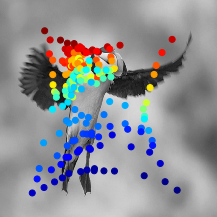}}\hfill	
	\subfigure[(c)]
	{\includegraphics[width=0.163\linewidth]{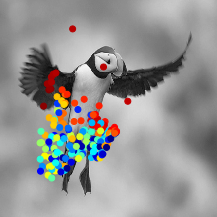}}\hfill	
	\subfigure[(d)]
	{\includegraphics[width=0.163\linewidth]{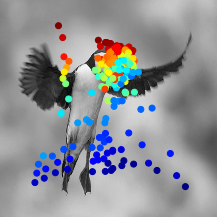}}\hfill	
	\subfigure[(e)]
	{\includegraphics[width=0.163\linewidth]{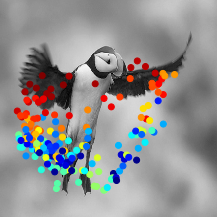}}\hfill	
	\subfigure[(f)]
	{\includegraphics[width=0.163\linewidth]{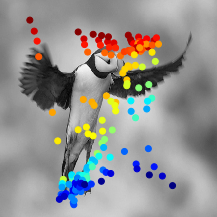}}\hfill\\	
	\vspace{-10pt}
	\caption{Visualization of VTN: (a) input image and target coordinates for warping an intermediate CNN feature map, (b) source coordinates obtained using STNs~\cite{Jaderberg15} (top) and SSN~\cite{Recasens18} (bottom), and (c), (d), (e), and (f) four feature channels and samplers in VTN. Note that the colors in the warping fields represent the corresponding target coordinates. Unlike STNs~\cite{Jaderberg15} that applies the same warping field across all the feature channels, VTN maps the individual channels independently to the canonical configuration, by localizing different semantic parts in different channels.}\label{img:1}\vspace{-10pt}
\end{figure}

In this paper, we address this by introducing a learnable module, the volumetric transformer network (VTN), that predicts channel-wise warping fields. As illustrated by~\figref{img:1}, this allows us to correctly account for the different transformations of different semantic parts by reconfiguring the intermediate features of a CNN spatially and channel-wisely.
%, to boost the representation of an intermediate CNN feature map by reconfiguring the features spatially and channel-wisely, . 
%Based on the observation that each feature channel encodes individual semantic meaning separately, VTNs estimates spatial and channel-wise warping fields at each spatial and channel-wise point that localize the most discriminative parts to boost a given task, unlike STNs~\cite{Jaderberg15} that only applies identical warping fields to all the channels. We cast this as a dense prediction of the warping field at whole points in an intermediate feature. 
To achieve this while nonetheless accounting for the dependencies between the different semantic parts, 
%\ms{e.g., the position of the bird's wings are influenced by that of its body,} 
we introduce an encoder-decoder network that lets information flow across the original feature channels. Specifically, our encoder relies on a channel-squeeze module that aggregates information across the channels, while our decoder uses a channel-expansion component that distributes it back to the original features.

%While one could think of doing so by exploiting 3D convolutions~\cite{Tran15} across the spatial and channel axes, this would fail to account for the fact that the different feature channels are obtained with different parameters, rendering meaningless the notion of neighborhood \ms{(in depth)} used in 3D convolutions. To overcome this, we use an encoder-decoder network whose encoder where we decompose the input feature map across the channel dimension and exploit channel squeeze and expansion modules to aggregate and distribute information across the channels within the encoder and decoder, respectively.

%but the convolution operators process a local neighborhood across the channel cannot be used, since the feature channels are not locally similar rather independent~\cite{He16,Jeong19,Zhang19}. To solve this, we separate spatial- and channel-convolution within an encoder-decoder network, where we exploit channel squeeze and expansion layers to aggregate and distribute information across channels within the encoder and decoder networks, respectively. 

As shown in previous works~\cite{Jaderberg15,Yi16,Lin17inv,Esteves18}, training a localization network to achieve spatial invariance is challenging, and most methods~\cite{Jaderberg15,Yi16,Lin17inv,Esteves18} rely on indirect supervision via a task-dependent loss function, as supervision for the warping fields is typically unavailable. This, however, does not guarantee that the warped features are consistent across different object instances. To improve the localization ability of the predicted warping fields, we further introduce a loss function defined between the warped features of pairs of instances, so as to encourage similarity between the representation of same-class instances while pushing that of different-class instances apart.
%such that the correct warping fields are identified by the matching score. \MS{I do not understand this.}

Our experiments on fine-grained image recognition~\cite{WelinderEtal2010,Krause15,Maji13,Horn18} and instance-level image retrieval~\cite{Radenovic18}, performed using several backbone networks and pooling methods, evidence that our VTNs consistently boost the features' representation power and consequently the networks' accuracy. \vspace{-5pt}

\section{Related Work}
\noindent{{\bf Attention mechanisms.}} 
As argued in~\cite{Zhu19}, spatial deformation modeling methods~\cite{Jaderberg15,Lin17inv,Dai17,Recasens18}, including VTNs, can be viewed as hard attention mechanisms, in that they localize and attend to the discriminative image parts. 
Attention mechanisms in neural networks have quickly gained popularity in diverse computer vision and natural language processing tasks, such as relational reasoning among objects~\cite{Battaglia16,Santoro17}, image captioning~\cite{Xu15}, neural machine translation~\cite{Bahdanau15,Vaswani17}, image generation~\cite{Xu18,Zhang19}, and image recognition~\cite{Hu18,Wang18}. They draw their inspiration from the human visual system, which understands a scene by capturing a sequence of partial glimpses and selectively focusing on salient regions~\cite{Itti98,Larochelle10}. 

Unlike methods that consider spatial attention~\cite{Gregor15,Xu15,Noh17,Woo18}, some works~\cite{Wang17,Zhang18,Hu18,Fu19} have attempted to extract channel-wise attention based on the observation that different feature channels can encode different semantic concepts~\cite{Ren15,Bau17,Garcia18}, so as to capture the correlations among those concepts. In those cases, however, spatial attention was ignored. While some methods~\cite{Chen17,Woo18} have tried to learn spatial and channel-wise attention simultaneously, they only predict a fixed spatial attention with different channel attentions. More importantly, attention mechanisms have limited ability to handle spatial variations due to, e.g., scale, pose and viewpoint changes, or part deformations~\cite{Felzenszwalb10,Jaderberg15,Dai17}.
\vspace{+10pt}

\noindent{{\bf Spatial invariance.}} 
Recent work on spatial deformation modeling seeks to spatially warp the features to a canonical configuration so as to facilitate recognition~\cite{Jaderberg15,Lin17inv,Dai17,Esteves18,Recasens18}. STNs~\cite{Jaderberg15} explicitly allow the spatial manipulation of feature maps within the network while attending to the discriminative parts. Their success inspired many variants that use, e.g., a recurrent formalism~\cite{Lin17inv}, polar transformations~\cite{Esteves18}, deformable convolutional kernels~\cite{Dai17}, and attention based warping~\cite{Recasens18,Zheng19}. These methods typically employ an additional network, called localization network, to predict a warping field, which is then applied to all the feature channels identically. Conceptually, this corresponds to using hard attention~\cite{Gregor15,Xu15,Noh17,Woo18}, but it improves spatial invariance. While effective, this approach concentrates on finding the regions that are most discriminative across all the feature channels. To overcome this, some methods use multi-branches~\cite{Jaderberg15,Zheng17}, coarse-to-fine schemes~\cite{Fu17}, and recurrent formulations~\cite{Li17}, but they remain limited to considering a pre-defined number of discriminative parts, which restricts their effectiveness and flexibility.
\vspace{+10pt}

\noindent{{\bf Fine-grained image recognition.}} 
To learn discriminative feature representations of object parts, conventional methods first localize these parts and then classify the whole image based on the discriminative regions. These two-step methods~\cite{Berg14,Huang16} typically require bounding box or keypoint annotations of objects or parts, which are hard to collect. To alleviate this, recent methods aim to automatically localize the discriminative object parts using an attention mechanism~\cite{Fu17,Zheng17,Li17,Sun18,Rodriguez18,Recasens18,Chen19,Zheng19} in an unsupervised manner, without part annotations. However, these methods do not search for semantic part representations in the individual feature channels, which limits their ability to boost the feature representation power. Recently, Chen et al.~\cite{Chen19} proposed a destruction and construction learning strategy that injects more discriminative local details into the classification network. However, the problem of explicitly processing the individual feature channels remains untouched.
\vspace{+10pt}

\noindent{{\bf Instance-level image retrieval.}} 
While image retrieval was traditionally tackled using local invariant features~\cite{Lowe04,Mikolajczyk05} or bag-of-words (BoW) models~\cite{Sivic03,Arandjelovic13}, recent methods use deep CNNs~\cite{Babenko15,Tolias16,Kalantidis16,Tolias16,Noh17,Radenovi19tpami} due to their better representation ability. In this context, the main focus has been on
improving the feature representation power of pretrained backbone networks~\cite{Krizhevsky12,Simonyan15,He16}, typically by designing pooling mechanisms to construct a global feature, such as max-pooling (MAC)~\cite{Tolias16}, sum-pooling (SPoC)~\cite{Babenko15}, weighted sum-pooling (CroW)~\cite{Kalantidis16}, regional max-pooling (R-MAC)~\cite{Tolias16}, and generalized mean-pooling (GeM)~\cite{Radenovi19tpami}. These pooling strategies, however, do not explicitly leverage the discriminative parts, and neither do the methods~\cite{Gordo17,Radenovi19tpami} that have tried to fine-tune the pretrained backbone networks~\cite{Krizhevsky12,Simonyan15,He16}. While the approach of~\cite{Noh17} does, by learning spatial attention, it ignores the channel-wise variations. Taking such variations into account is the topic of this paper.\vspace{-5pt}

\section{Volumetric Transformer Networks}
\subsection{Preliminaries}
Let us denote an intermediate CNN feature map as ${U} \in \mathbb{R}^{H\times W\times K}$, with height $H$, width $W$, and
$K$ channels. To attend to the discriminative object parts and reduce the inter-instance spatial variations in the feature map, recent works~\cite{Jaderberg15,Yi16,Lin17inv,Esteves18} predict a warping field to transform the features to a canonical pose. 
This is achieved via a module that takes ${U}$ as input and outputs the parameters defining a warping field ${G} \in \mathbb{R}^{H\times W\times 2}$ to be applied to ${U}$. The representation in the canonical pose is then obtained via a feature sampling mechanism, which, for every pixel $i$ in the output representation, produces a warped feature such that ${V(i)}=U(i+G(i))$. As argued above, while this reduces spatial variations and lets the network focus on discriminative image regions,
%object parts, 
the same warping field is applied across all the channels, without considering the different semantic meanings of these individual channels.
%, thus limiting the features' representation ability. 
Moreover, this does not explicitly constrain the warped features of different instances of the same class to be consistent. \vspace{-5pt}
\begin{figure*}[t!]
	\centering
	\renewcommand{\thesubfigure}{}
	\subfigure[(a)]
	{\includegraphics[width=0.248\linewidth]{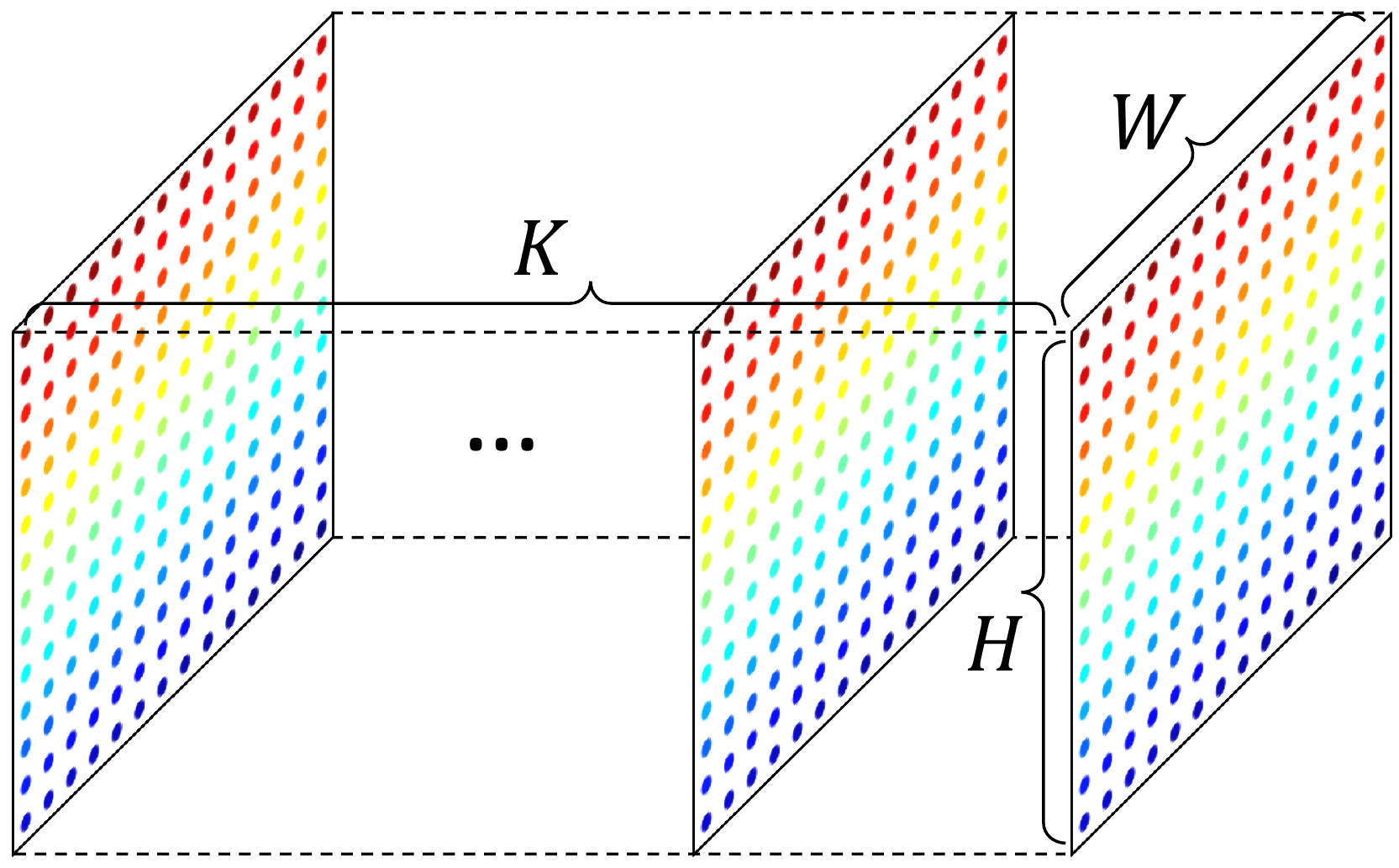}}\hfill
	\subfigure[(b)]
	{\includegraphics[width=0.248\linewidth]{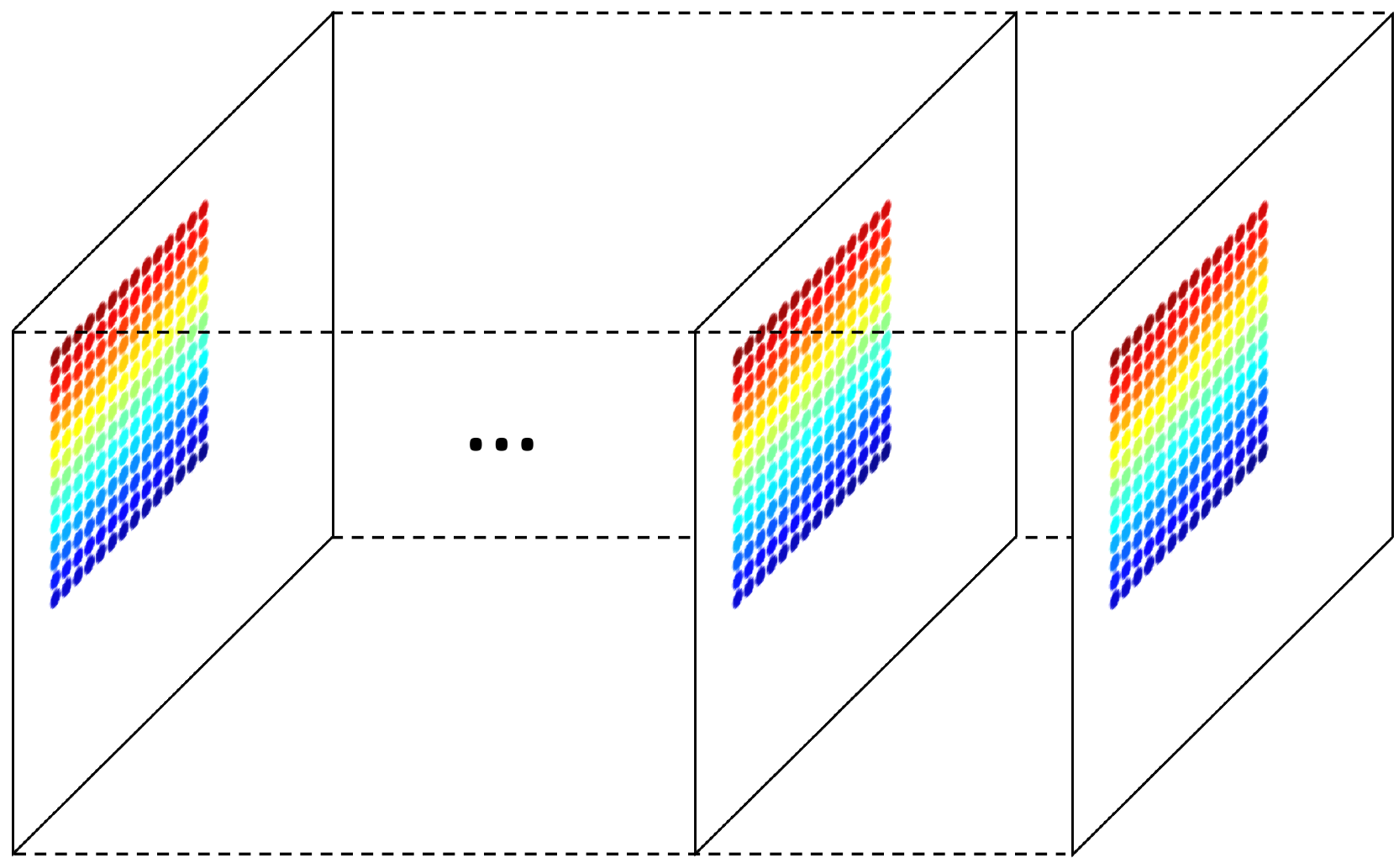}}\hfill
	\subfigure[(c)]
	{\includegraphics[width=0.248\linewidth]{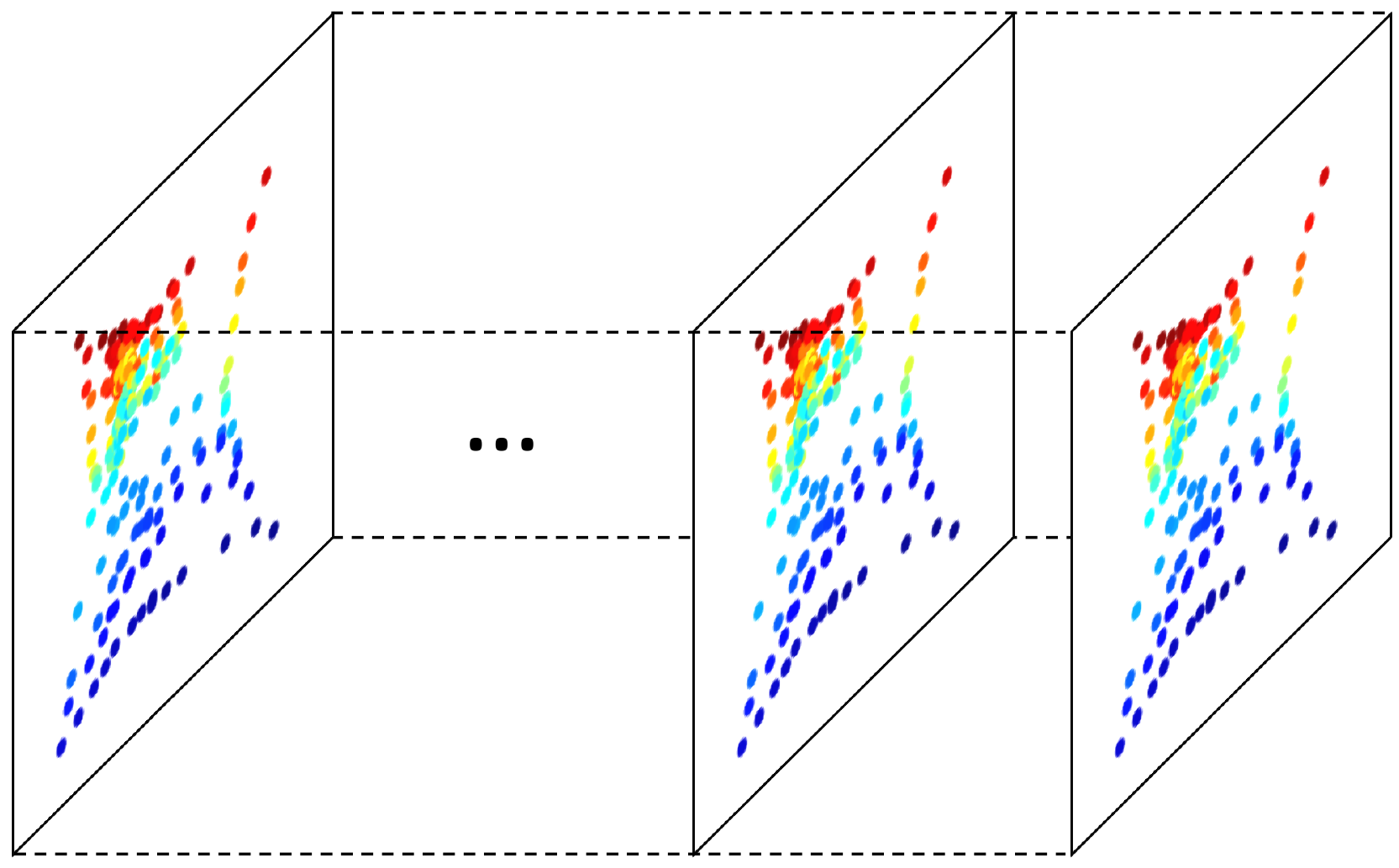}}\hfill
	\subfigure[(d)]
	{\includegraphics[width=0.248\linewidth]{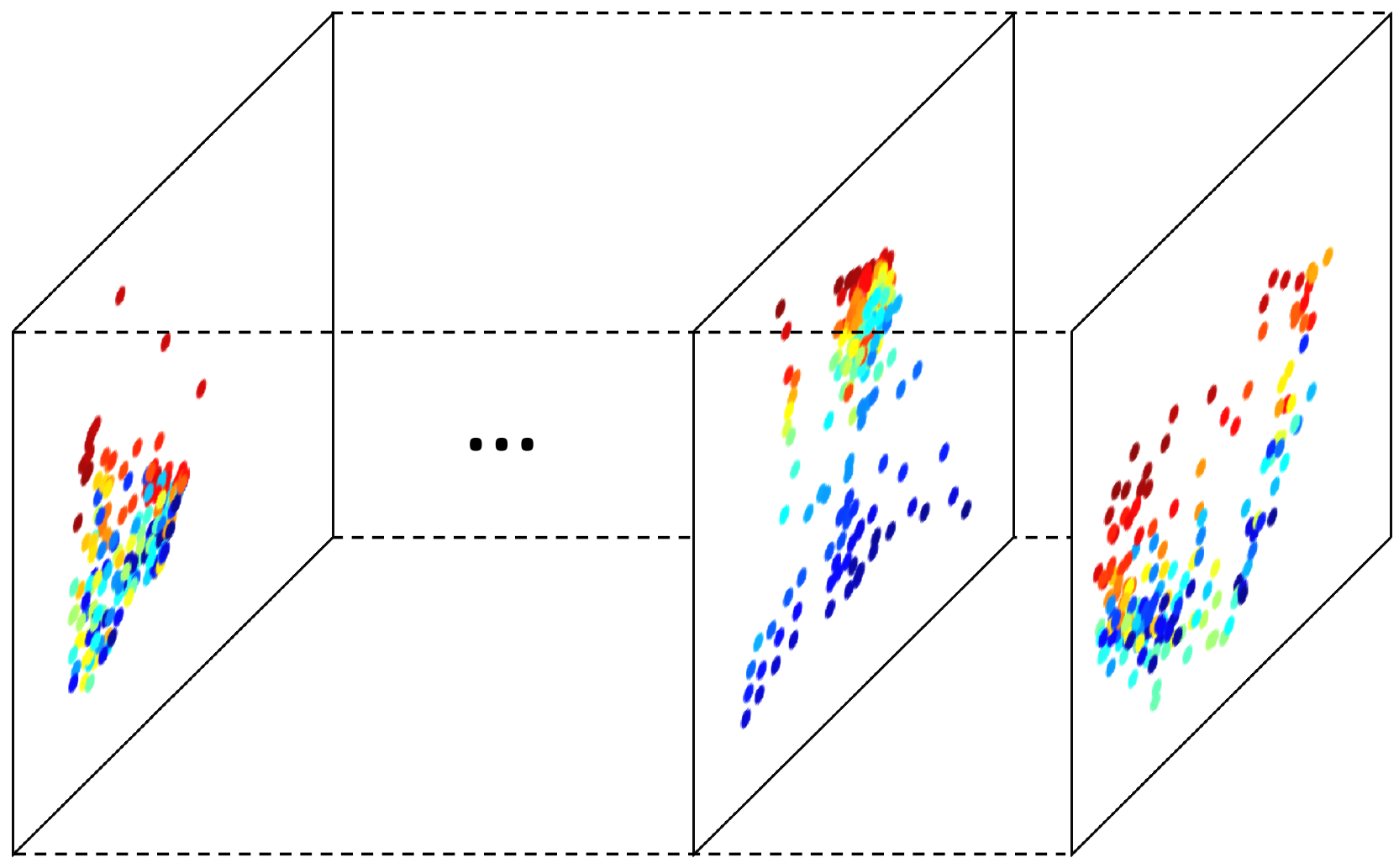}}\hfill\\
	\vspace{-10pt}
	\caption{Intuition of VTNs: (a) target coordinates for warping an intermediate CNN feature map and source coordinates obtained using (b) STNs~\cite{Jaderberg15}, (c) SSN~\cite{Recasens18}, and (d) VTNs, which predict channel-wise warping fields.
	}\label{img:2}\vspace{-10pt}
\end{figure*} 

\subsection{Motivation and Overview}
By contrast, our volumetric transformer network (VTN), which we introduce in the remainder of this section, encodes the observation that each channel in an intermediate CNN feature map acts as a pattern detector, i.e., high-level channels detect high-level semantic patterns, such as parts and objects~\cite{Bau17,Chen17,Garcia18,Woo18}, and, because these patterns can undergo different transformations, one should separately attend to the discriminative parts represented by the individual channels to more effectively warp them to the canonical pose.
%to increase the network's representation power and allow it to encode stronger correlations between semantic patterns. 
To achieve this, unlike existing spatial deformation modeling  methods~\cite{Jaderberg15,Lin17inv,Dai17,Recasens18}, which apply the same warping field to all the feature channels, as in~\figref{img:2}(b), our VTN predicts channel-wise warping fields, shown in~\figref{img:2}(d).
%, which enables a non-linear feature warping to reconfigurate the feature and to boost the following tasks. 

Concretely, a VTN produces a warping field ${G}_c\in \mathbb{R}^{H\times W\times 2}$ for each channel $c$. Rather than estimating the warping field of each channel independently, to account for dependencies between the different semantic parts, we design two modules, the channel squeeze and expansion modules, that let information flow across the channels.
%we exploit channel squeeze and expansion modules within an encoder-decoder network to account for inter-channel relationships. 
Furthermore, to improve the computational efficiency and localization accuracy, we build a group sampling and normalization module, and a transformation probability inference module at the first and last layer of VTN, respectively. To train the network, instead of relying solely on a task-dependent loss function as in~\cite{Jaderberg15,Lin17inv,Dai17,Recasens18}, which may yield poorly-localized warping fields, we further introduce a loss function based on the distance between the warped features of pairs of instances, thus explicitly encouraging the warped features to be consistent across different instances of the same class.\vspace{-5pt}
%to leverage an image pair based on the intuition that warped source and target features should be matched.

\subsection{Volumetric Transformation Estimator}
Perhaps the most straightforward way to estimate channel-wise warping fields is to utilize convolutional layers that take the feature map ${U}$ as input and output the warping fields $G = \{G_c\} \in \mathbb{R}^{H \times W \times 2 \times K}$. This strategy, however,
uses separate convolution kernels for each warping field $G_c$, which might be subject to overfitting because of the large number of parameters involved.
%hard to train and cannot consider inter-channel relationship. 
%\MS{I disagree that it cannot consider inter-channel relationship: Every warping field would be predicted by using all the channels as input.}
As an alternative, one can predict each warping field $G_c$ independently, by taking only the corresponding feature channel ${U}_c \in \mathbb{R}^{H \times W\times1}$ as input.
%with spatially shared parameters $\mathbf{W}$ such that $G_c=\mathcal{F}({X}_c;\mathbf{W})$. 
%As it is a pixel-level prediction problem, we can use any kind of network architectures for dense correspondence such as encoder-decoder~\cite{Hinton06} or U-Net~\cite{Ronneberger15}. 
This, however, would fail to account for the inter-channel relationships, and may be vulnerable to outlier channels that, on their own, contain uninformative features but can yet be supported by other channels~\cite{Garcia18,Zhang19,Jeong19}. 
%Moreover, the computational time linearly increases  with the number of channels. 

To alleviate these limitations, we introduce the channel squeeze and expansion modules, which yield a trade-off between the two extreme solutions discussed above. We first decompose the input feature map across the channel dimension, and apply a shared convolution to each of the $K$ channels. We then combine the original feature channels into $K'$ new channels by a channel-squeeze module, parameterized by a learned matrix $W_\mathrm{cs}$, in the encoder and expand these squeezed feature channels into $K$ channels by a channel-expansion module, parameterized by a learned matrix $W_\mathrm{ce}$, in the decoder.

\begin{figure*}[t]
	\centering
	\renewcommand{\thesubfigure}{}
	\subfigure[]
	{\includegraphics[width=1\linewidth]{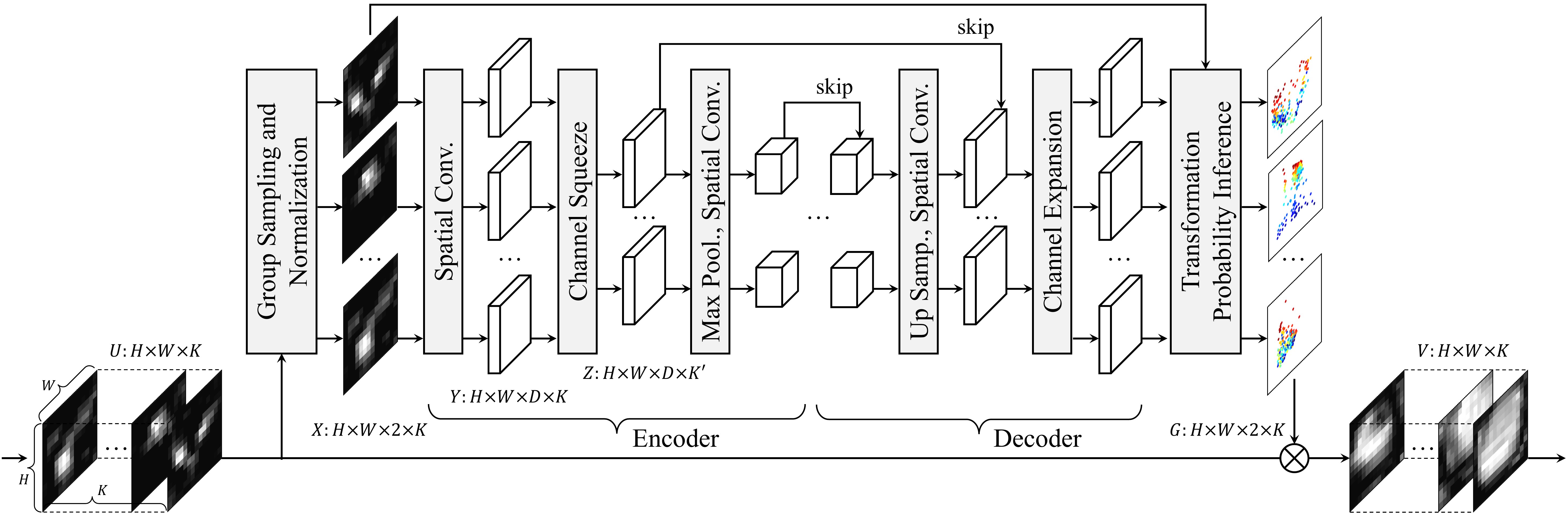}}\\
	\vspace{-20pt}
	\caption{VTN architecture. A VTN consists of a group sampling and normalization module, sequential spatial convolutions, channel squeeze and expansion modules, and a transformation probability inference module.}\label{img:4}\vspace{-10pt}
\end{figure*}

Formally, as depicted by \figref{img:4}, let us define an intermediate CNN feature map after a forward pass through an encoder as $Y=\mathcal{F}({U};{W}_\mathrm{s})\in\mathbb{R}^{H\times W\times D \times K}$, where each feature channel is processed independently with spatial convolution parameters ${W}_\mathrm{s}$ shared across the channels, introducing an additional dimension of size $D$.
%\MS{What do you mean by spatial convolution parameters? Do you mean that each input channel is processed independently and the parameters are shared across the channels? Maybe indicating the dimension of $\mathbf{W}_\mathrm{s}$ would help.} 
We introduce a channel squeeze module, with parameters ${W}_\mathrm{cs} \in \mathbb{R}^{K \times K'}$, $K'<K$, applied to the
reshaped $Y\in\mathbb{R}^{HWD \times K}$,
%\MS{The dimensions here don't really match. Is it a 1x1 convolution? How about the dimension $D$?}, 
whose role is to aggregate the intermediate features so as to output $Z = \mathcal{F}(Y;{W}_\mathrm{cs})\in\mathbb{R}^{HWD\times K'}$, which can also be reshaped to $\mathbb{R}^{H\times W\times D\times K'}$. In short, this operation allows the network to learn how to combine the initial $K$ channels so as to leverage the inter-channel relationships while keeping the number of trainable parameters reasonable.
%The networks learn the contribution of $Y_c$ to $Y'_c$. By using the channel squeeze networks, the networks not only leverage inter-channel relationship but also reduce the computation complexity. 
We then incorporate a channel expansion module, with parameters ${W}_\mathrm{ce} \in \mathbb{R}^{K' \times K}$, which performs the reverse operation, thereby enlarging the feature map $Z\in\mathbb{R}^{H\times W\times D\times K'}$ back into a representation with $K$ channels. This is achieved through a decoder. 
%\MS{As before, write down all dimensions}.

We exploit sequential spatial convolution and channel squeeze modules in the encoder, and sequential spatial convolution and channel expansion modules in the decoder. In our experiments, the volumetric transformation estimator consists of an encoder with 4 spatial convolution and channel squeeze modules followed by max-pooling, and a decoder with 4 spatial convolution and channel expansion modules followed by upsampling. Each convolution module follows the architecture Convolution-BatchNorm-ReLU~\cite{Ioffe15}. \vspace{+10pt}

\noindent{{\bf Grouping the channels.}} 
%Since each feature slice ${U}_c$ is built using an independent kernel, there exist response variations among them and one filter alone might be insufficient~\cite{Garcia18,Jeong19,Zhang19}. 
In practice, most state-of-the-art networks~\cite{Simonyan15,He16,huang2017densely} extract high-dimensional features, and thus processing all the initial feature channels as described above can be computationally prohibitive. To overcome this, we propose a group sampling and normalization module inspired by group normalization~\cite{Wu18} and attention mechanisms~\cite{Woo18}. Concretely, a group sampling and normalization module takes the feature map ${U}$ as input and separates it into $C$ groups following the sequential order of the channels.
%, where each group is sampled in a sequential order along the channel axis. %\MS{How is the grouping done?} 
We then aggregate the features $U_c$ in each group $c\in\{1,\dots,C\}$ by using two pooling operations: ${U}^\mathrm{max}_c \in \mathbb{R}^{H\times W\times 1}$ and ${U}^\mathrm{avg}_c \in \mathbb{R}^{H\times W\times 1}$, and concatenate them as ${X}_c \in \mathbb{R}^{H\times W\times 2}$, followed by group normalization without per-channel linear transform~\cite{Wu18}. We then take the resulting $X = \{{X}_c\}\in \mathbb{R}^{H \times W \times 2 \times K}$ as input to the volumetric transformation estimator described above, instead of $U$. \vspace{+10pt}
\begin{figure*}[t]
	\centering
	\renewcommand{\thesubfigure}{}
	\subfigure[(a)]
	{\includegraphics[width=0.15\linewidth]{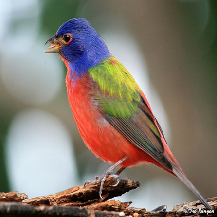}}
	\subfigure[(b)]
	{\includegraphics[width=0.15\linewidth]{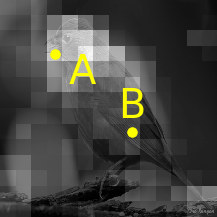}}
	\subfigure[(c)]
	{\includegraphics[width=0.15\linewidth]{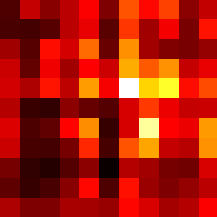}}
	\subfigure[(d)]
	{\includegraphics[width=0.15\linewidth]{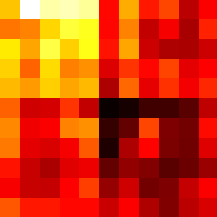}}\\
	\vspace{-10pt}
	\caption{Visualization of the learned probability of transformation candidates: (a) input image, (b) arbitrary feature channel, and (c) and (d) probabilities of points A and B in (b), respectively. VTNs estimate the probability of transformation candidates, instead of directly estimating the warping fields.}\label{img:5}\vspace{-10pt}
\end{figure*}

\noindent{{\bf Probabilistic transformation modeling.}} 
Unlike existing spatial deformation modeling methods~\cite{Jaderberg15,Lin17inv} that rely on parametric models, e.g., affine transformation, VTNs estimate non-parametric warping fields, thus having more flexibility. However, regressing the warping fields directly may perform poorly because the mapping from the features to the warping fields adds unnecessary learning complexity. To alleviate this, inspired by~\cite{Tompson15,Sun19}, we design a probabilistic transformation inference module that predicts probabilities for warping candidates, instead of directly estimating the warping field. Specifically, we predict the probability $P_c(i,j)$ of each candidate $j \in {N}_i$ at each pixel $i$, and compute
the warping field ${G}_c$ by aggregating these probabilities
%using a soft-argmin 
as
\begin{equation}
{G}_c(i) = \sum\limits_{j \in {N}_i} P_c(i,j) (j-i).
\end{equation}
Furthermore, instead of predicting the probability $P_c(i,j)$ directly, we compute a residual probability and then use a softmax layer such that 
\begin{equation}
P_c(i,j) = \Psi\left(\left({{U}^\mathrm{max}_c(j)+{U}^\mathrm{avg}_c(j)+E_c(i,j)}\right)/{\beta}\right),
\end{equation}
where $E_c \in \mathbb{R}^{H\times W\times |{N}_i|}$ is the output of the volumetric transformation estimator whose the size depends on the number of candidates $|{N}_i|$. Note that $E_c(i,j)$ is a scalar because $i$ denotes a spatial point over $H\times W$ and $j$ indexes a point among all candidates. $\Psi(\cdot)$ is the softmax operator and $\beta$ is a parameter adjusting the sharpness of the softmax output. At initialization, the network parameters are set to predict zeros, i.e., $E_c(i,j)=0$, thus the warping fields are determined by candidate feature responses $U^\mathrm{max}_c+U^\mathrm{avg}_c$, which provide good starting points. As training progresses, the network provides increasingly regularized warping fields. This is used as the last layer of the VTN. \figref{img:5} visualizes the learned probability of some transformation candidates.\vspace{-5pt}
\begin{SCfigure}[][t]
	\includegraphics[width=0.5\linewidth]{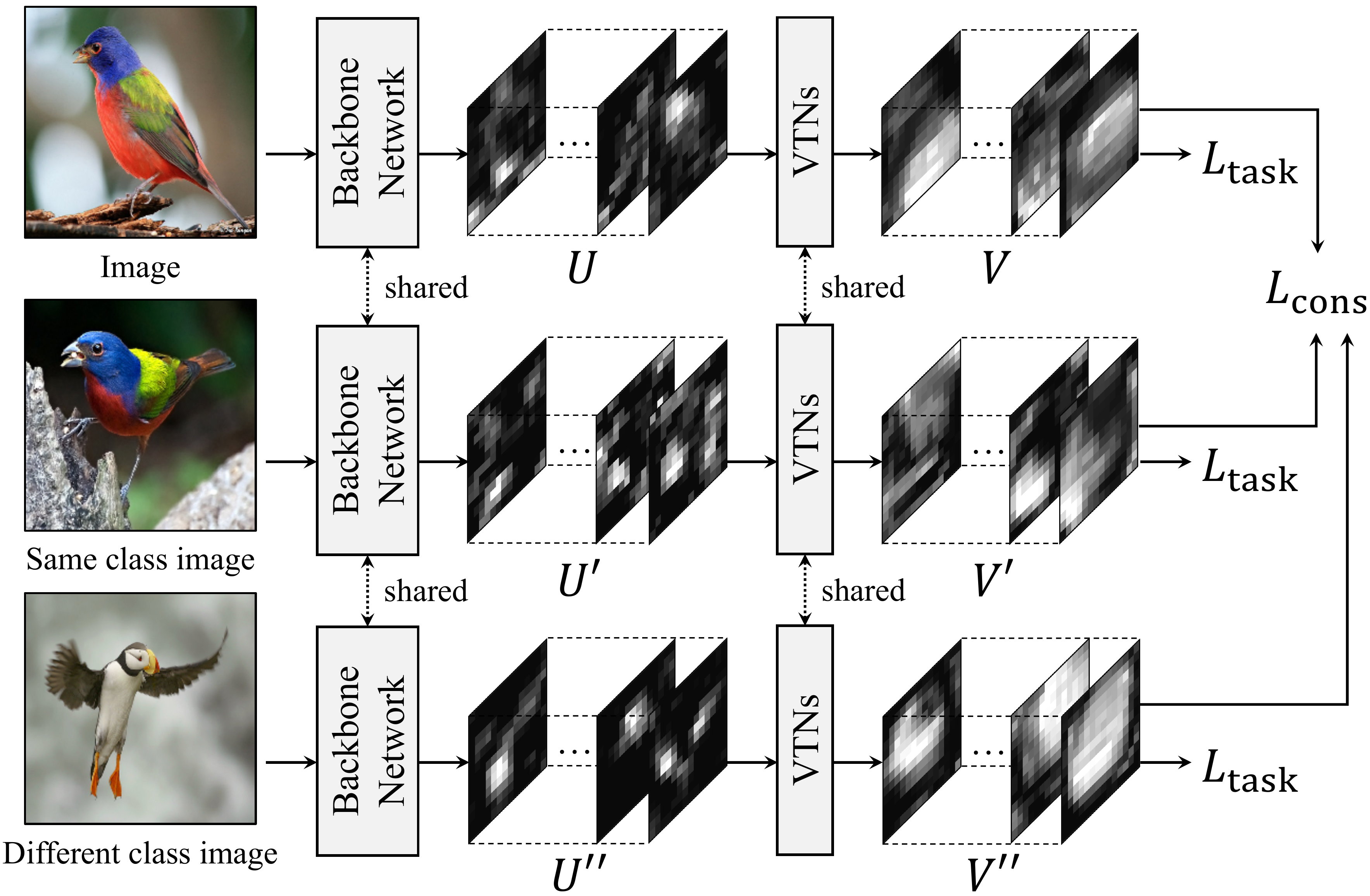}
	\caption{Illustration of training VTNs using our consistency loss function. By simultaneously exploiting a sample from the same class and another sample from a different class, our consistency loss function improves the localization ability and the discriminative power of the intermediate features.}\label{img:7}\vspace{-10pt}
\end{SCfigure}

\subsection{Loss Function}
Similarly to existing deformation modeling methods~\cite{Jaderberg15,Yi16,Lin17inv,Esteves18}, our network can  be learned using only the final task-dependent loss function $\mathcal{L}_\mathrm{task}$, without using ground-truth warping fields, since all modules are differentiable. This, however, does not explicitly constrain the warped features obtained from the  predicted warping fields to be consistent across different object instances of the same class. To overcome this, we draw our inspiration from semantic correspondence works~\cite{Rocco18,Kim19}, and introduce an additional loss function modeling the intuition that the warped features of two instances of the same class should match and be similar. The simplest approach to encoding this consists of using a square loss between such features, which yields
\begin{equation}
\mathcal{L}
= \sum\limits_{i} {\|V({i})-V'({i})\|}^2, 
\end{equation}
where $V$ and $V'$ are the warped feature maps of two instances of the same class. 
%\MS{Then, isn't $V_{i,c}$ a scalar value? If so, then we should not use the norm symbol. If not, what is its dimension?}
%By minimizing the loss function $\mathcal{L}_\mathrm{cons}$ for randomly sampled feature pairs $U$ and $U'$ from the same category, the networks learn consistently warped features $V$ and $V'$ with corresponding warping fields $G$ and $G'$, respectively, in a manner that same points represent same feature responses. 
Minimizing this loss function, however, can induce erroneous solutions, such as constant features at all the pixels. To avoid such trivial solutions, we use a triplet loss function~\cite{Simo-Serra15b,Yi16} simultaneously exploiting a sample $V'$ from the same class as $V$ and another sample $V''$ from a different class. We then express our loss as 
\begin{equation}
\mathcal{L}_\mathrm{cons} = \sum\limits_{i}
\left[\|V({i})-V'(i)\|^2-\|V({i})-V''(i)\|^2+\alpha \right]_{+},
\end{equation}
where $\alpha>0$ is a threshold parameter and $\left[\cdot \right]_{+} = \mathrm{max}(\cdot,0)$. 
Our loss function jointly encourages the instances' features from the
same class to be similar, and the instances' features from different
classes to be dissimilar. Together,
this helps to improve the features' discriminative power, which is
superior to relying solely on a task-dependent loss function, as in
previous methods~\cite{Jaderberg15,Yi16,Lin17inv,Esteves18}.
%\figref{img:6} visualizes the effect of the proposed loss function.
%\MS{Why was this not possible? Was it truly not possible, or is it just that people did not do it?} 
Note that our approach constitutes the first attempt at learning warping fields that generate consistent warped features across object instances.
%using the weak supervision in the form of image pairs. 
To train our VTNs, we then use the total loss
$\mathcal{L}_\mathrm{total}=\mathcal{L}_\mathrm{task}+\lambda\mathcal{L}_\mathrm{cons}$ with balancing parameter $\lambda$. \figref{img:7} depicts the training procedure of VTNs.\vspace{-5pt}

\subsection{Implementation and Training Details}
We implemented VTNs using the {\ttfamily Pytorch} library~\cite{paszke2017automatic}. In our experiments, we use VGGNet~\cite{Simonyan15} and ResNet~\cite{He16} backbones pretrained on ImageNet~\cite{imagenet_cvpr09}. 
We build VTNs on the last convolution layers of each network.
For fine-grained image recognition, we replace the 1000-way softmax layer with a $k$-way softmax layer, where $k$ is number of classes in the dataset~\cite{WelinderEtal2010,Krause15,Maji13,Horn18}, and fine-tune the networks on the dataset. The input images were resized to a fixed resolution of $512\times512$ and randomly cropped to $448\times448$. We apply random rotations and random horizontal flips for data augmentation. 
For instance-level image retrieval, we utilize the last convolutional features after the VTNs as global representation. To train VTNs, we follow the experimental protocols of~\cite{Radenovi19tpami,Gordo17}. We set the hyper-parameters by cross-validation on CUB-Birds~\cite{WelinderEtal2010}, and then used the same values for all experiments. We set the size of the transformation candidates $|{N}_i|=11\times11$, the parameter $\beta=10$, the number of groups $C=32$, and the balancing parameter $\lambda=1$. 
We also set the threshold parameter $\alpha=20$ for VGGNet~\cite{Simonyan15} and $\alpha=30$ for ResNet~\cite{He16}, respectively, because they have different feature distributions. The source code is available online at our project webpage: {\ttfamily http://github.com/seungryong/VTNs/}. \vspace{-5pt}
\begin{table}[t]
	\begin{center}
		\begin{tabular}{ >{\raggedright}m{0.3\linewidth}
				>{\centering}m{0.2\linewidth} >{\centering}m{0.10\linewidth}
				>{\centering}m{0.10\linewidth} >{\centering}m{0.10\linewidth}}
			\hlinewd{0.8pt}
			Methods &Backbone &\cite{WelinderEtal2010} &\cite{Krause15} &\cite{Maji13} \tabularnewline
			\hline
			\hline
			\multirow{2}{*}{Base} 
			&VGG-19 &71.4 &68.7 &80.7\tabularnewline
			&ResNet-50 &74.6 &70.4 &82.1\tabularnewline
			\hline
			\multirow{2}{*}{Def-Conv~\cite{Dai17}} 
			&VGG-19 &74.2 &70.1 &82.6\tabularnewline
			&ResNet-50 &76.7 &72.1 &83.7\tabularnewline
			\multirow{2}{*}{STNs~\cite{Jaderberg15}} 
			&VGG-19 &72.1 &69.2 &81.1\tabularnewline
			&ResNet-50 &76.5 &71.0 &81.2\tabularnewline	
			\multirow{2}{*}{SSN~\cite{Recasens18}} 
			&VGG-19 &75.1 &72.7 &84.6\tabularnewline
			&ResNet-50 &77.7 &74.8 &83.1\tabularnewline
			\multirow{2}{*}{ASN~\cite{Zheng19}} 
			&VGG-19 &76.2 &74.1 &82.4\tabularnewline
			&ResNet-50 &78.9 &75.2 &85.7\tabularnewline
			\hline
			\multirow{2}{*}{{\bf VTNs} wo/${W}_\mathrm{cs},{W}_\mathrm{ce}$} 
			&VGG-19 &77.8 &78.6 &86.1\tabularnewline
			&ResNet-50 &80.1 &81.4 &86.9\tabularnewline
			\multirow{2}{*}{{\bf VTNs} wo/Group} 
			&VGG-19 &76.3 &76.1 &84.4\tabularnewline
			&ResNet-50 &77.2 &79.1 &82.4\tabularnewline
			\multirow{2}{*}{{\bf VTNs} wo/T-Probability} 
			&VGG-19 &78.1 &79.7 &84.9\tabularnewline
			&ResNet-50 &79.0 &80.4 &85.1\tabularnewline
			\multirow{2}{*}{{\bf VTNs} wo/$\mathcal{L}_\mathrm{cons}$} 
			&VGG-19 &79.2 &80.2 &87.1\tabularnewline
			&ResNet-50 &82.4 &82.1 &84.9\tabularnewline
			\multirow{2}{*}{{\bf VTNs}}
			&VGG-19 &80.4 &81.9 &87.4\tabularnewline
			&ResNet-50 &\underline{\bf{83.1}} &\underline{\bf{82.7}} &\underline{\bf{89.2}}\tabularnewline
			\hlinewd{0.8pt}
		\end{tabular}
	\end{center}
	\vspace{-5pt}
	\caption{Accuracy of VTNs compared to spatial deformation modeling methods on fine-grained image recognition benchmarks (CUB-Birds~\cite{WelinderEtal2010}, Stanford-Cars~\cite{Krause15}, and FGVC-Aircraft~\cite{Maji13}).}\label{tab:1}\vspace{-20pt}
\end{table}
\begin{figure*}[t!]
	\centering
	\renewcommand{\thesubfigure}{}
	\subfigure[]
	{\includegraphics[width=0.122\linewidth]{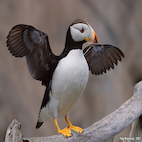}}\hfill
	\subfigure[]
	{\includegraphics[width=0.122\linewidth]{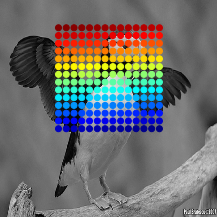}}\hfill
	\subfigure[]
	{\includegraphics[width=0.122\linewidth]{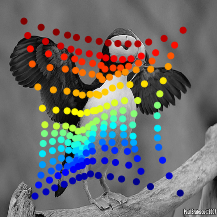}}\hfill
	\subfigure[]
	{\includegraphics[width=0.122\linewidth]{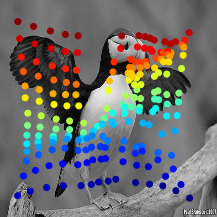}}\hfill
	\subfigure[]
	{\includegraphics[width=0.122\linewidth]{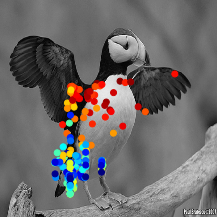}}\hfill
	\subfigure[]
	{\includegraphics[width=0.122\linewidth]{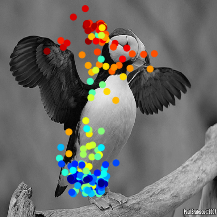}}\hfill
	\subfigure[]
	{\includegraphics[width=0.122\linewidth]{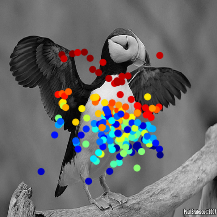}}\hfill
	\subfigure[]
	{\includegraphics[width=0.122\linewidth]{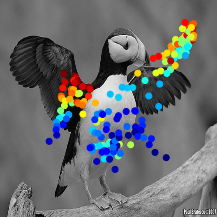}}\hfill\\
	\vspace{-20.5pt}
	\subfigure[(a)]
	{\includegraphics[width=0.122\linewidth]{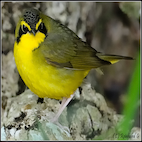}}\hfill
	\subfigure[(b)]
	{\includegraphics[width=0.122\linewidth]{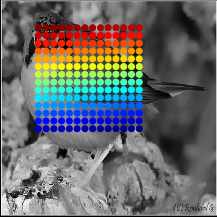}}\hfill
	\subfigure[(c)]
	{\includegraphics[width=0.122\linewidth]{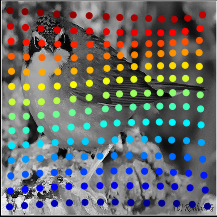}}\hfill
	\subfigure[(d)]
	{\includegraphics[width=0.122\linewidth]{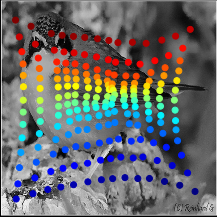}}\hfill
	\subfigure[(e)]
	{\includegraphics[width=0.122\linewidth]{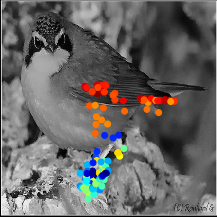}}\hfill
	\subfigure[(f)]
	{\includegraphics[width=0.122\linewidth]{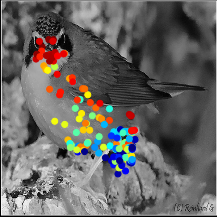}}\hfill
	\subfigure[(g)]
	{\includegraphics[width=0.122\linewidth]{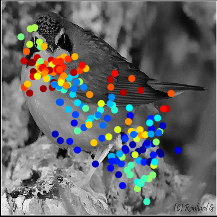}}\hfill
	\subfigure[(h)]
	{\includegraphics[width=0.122\linewidth]{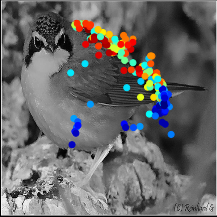}}\hfill\\
	\vspace{-10pt}
	\caption{Comparison of VTN warping fields with those of existing deformation modeling methods~\cite{Jaderberg15,Recasens18} on examples from CUB-Birds~\cite{WelinderEtal2010}: (a) input images and source coordinates obtained using (b) STNs~\cite{Jaderberg15}, (c) SSN~\cite{Recasens18}, (d) ASN~\cite{Zheng19}, and (e), (f), (g), and (h) four feature channel samplers in VTNs. Points with the same color in different images are projected to the same point in the canonical pose. This shows that VTNs not only localize different semantic parts in different channels but
		identify the same points across different images.}\label{img:8}\vspace{-5pt}
\end{figure*}

\section{Experiments}
\subsection{Experimental Setup}
In this section, we comprehensively analyze and evaluate VTNs on two tasks: fine-grained image recognition and instance-level image retrieval. First, we analyze the influence of the different components of VTNs compared to existing spatial deformation modeling methods~\cite{Jaderberg15,Dai17,Recasens18,Zheng19} and the impact of combining VTNs with different backbone networks~\cite{Simonyan15,He16} and second-order pooling strategies~\cite{Lin15,Gao16,Cui17,Li17,Li18}. Second, we compare VTNs with the state-of-the-art methods on fine-grained image recognition benchmarks~\cite{WelinderEtal2010,Krause15,Maji13,Horn18}. Finally, we evaluate them on instance-level image retrieval benchmarks~\cite{Radenovic18}.\vspace{-5pt}

\subsection{Fine-grained Image Recognition}
\noindent{{\bf Analysis of the VTN components.}} 
To validate the different components of our VTNs, we compare them with previous spatial deformation modeling methods, such as STNs~\cite{Jaderberg15}, deformable convolution (Def-Conv)~\cite{Dai17}, saliency-based sampler (SSN)~\cite{Recasens18}, and attention-based sampler (ASN)~\cite{Zheng19} on fine-grained image recognition benchmarks, such as CUB-Birds~\cite{WelinderEtal2010}, Stanford-Cars~\cite{Krause15}, and FGVC-Aircraft~\cite{Maji13}. For the comparison to be fair, we apply these methods at the same layer as ours, i.e., the last convolutional layer. In this set of experiments, we utilize VGGNet-19~\cite{Simonyan15} and ResNet-50~\cite{He16} as backbone networks.
As an ablation study, we evaluate VTNs without the channel squeeze and expansion modules, denoted by VTNs wo/${W}_\mathrm{cs},{W}_\mathrm{ce}$, without the group sampling and normalization module, denoted by VTNs wo/Group, and without the transformation probability inference module, denoted by VTNs wo/T-Probability. We further report the results of VTNs trained without our consistency loss function, denoted by VTNs wo/$\mathcal{L}_\mathrm{cons}$.

The results are provided in Table~\ref{tab:1} and \figref{img:8}. Note that all versions of our approach outperform the existing deformation modeling methods~\cite{Jaderberg15,Dai17,Recasens18,Zheng19}. Among these versions, considering jointly spatial and channel-wise deformation fields through our squeeze and expansion modules improves the results. 
The group sampling and normalization and transformation probability inference modules also boost the results. Using the consistency loss function $\mathcal{L}_\mathrm{cons}$ further yields higher accuracy by favoring learning a warping to a consistent canonical configuration of instances of the same class and improving the discriminative power of the intermediate features. \vspace{+10pt} 
\begin{table}[t]
	\begin{center}
		\begin{tabular}{ >{\raggedright}m{0.35\linewidth}
				>{\centering}m{0.10\linewidth}
				>{\centering}m{0.10\linewidth} >{\centering}m{0.10\linewidth}}
			\hlinewd{0.8pt}
			Methods &\cite{WelinderEtal2010} &\cite{Krause15} &\cite{Maji13} \tabularnewline
			\hline
			\hline
			Base &74.6 &70.4 &82.1\tabularnewline
			BP~\cite{Lin15}  &80.2 &81.5 &84.8\tabularnewline
			CBP~\cite{Gao16} &81.6 &81.6 &88.6\tabularnewline
			KP~\cite{Cui17} &83.2 &82.9 &89.9\tabularnewline
			MPN-COV~\cite{Li17} &84.2 &83.1 &89.7 \tabularnewline
			iSQRT-COV~\cite{Li18} &88.1 &90.0 &92.8\tabularnewline
			\hline
			Base+{\bf VTNs} &83.1 &82.7 &89.2\tabularnewline
			BP~\cite{Lin15}+{\bf VTNs} &84.9 &84.1 &90.6\tabularnewline
			CBP~\cite{Gao16}+{\bf VTNs} &85.2 &84.2 &91.2\tabularnewline
			KP~\cite{Cui17}+{\bf VTNs} &85.1 &83.2 &91.7\tabularnewline
			MPN-COV~\cite{Li17}+{\bf VTNs} &86.7 &88.1 &90.6\tabularnewline
			iSQRT-COV~\cite{Li18}+{\bf VTNs} &\underline{\bf{89.6}} &\underline{\bf{93.3}} &\underline{\bf{93.4}}\tabularnewline
			\hlinewd{0.8pt}
		\end{tabular}
	\end{center}
	\vspace{-5pt}
	\caption{Accuracy of VTNs incorporated with second-order pooling methods on fine-grained image recognition benchmarks (CUB-Birds~\cite{WelinderEtal2010}, Stanford-Cars~\cite{Krause15}, and FGVC-Aircraft~\cite{Maji13}).}\label{tab:2}\vspace{-20pt}
\end{table}
\begin{table}[!t]
	\begin{center}
		\begin{tabular}{ >{\raggedright}m{0.27\linewidth}
				>{\centering}m{0.24\linewidth} >{\centering}m{0.08\linewidth}
				>{\centering}m{0.08\linewidth} >{\centering}m{0.08\linewidth}}
			\hlinewd{0.8pt}
			Methods &Backbone &\cite{WelinderEtal2010} &\cite{Krause15} &\cite{Maji13} \tabularnewline
			\hline
			\hline
			RA-CNN~\cite{Fu17} &3$\times$VGG-19 &85.3 &92.5 &88.2\tabularnewline
			MA-CNN~\cite{Zheng17} &3$\times$VGG-19 &86.5 &92.8 &89.9\tabularnewline
			DFL-CNN~\cite{Wang18} &ResNet-50 &87.4 &93.1 &91.7\tabularnewline
			DT-RAM~\cite{Li17} &ResNet-50 &87.4 &93.1 &91.7\tabularnewline
			MAMC~\cite{Sun18} &ResNet-50 &86.5 &93.0 &92.9\tabularnewline
			NTSN~\cite{Sun18} &3$\times$ResNet-50 &87.5 &91.4 &93.1\tabularnewline
			\multirow{2}{*}{DCL~\cite{Chen19}} 
			&VGG-16 &86.9 &94.1 &91.2\tabularnewline
			&ResNet-50 &87.8 &94.5 &93.0\tabularnewline
			\multirow{2}{*}{TASN~\cite{Zheng19}} 
			&VGG-19&86.1 &93.2 &-\tabularnewline
			&ResNet-50 &87.9 &93.8 &-\tabularnewline
			\cite{Li18}+TASN~\cite{Zheng19}&ResNet-50 &89.1 &- &-\tabularnewline
			\hline
			DCL~\cite{Chen19}+{\bf VTNs}&ResNet-50 &89.2 &95.1 &93.4 \tabularnewline
			\cite{Li18}+{\bf VTNs}&ResNet-50 &89.6 &93.3 &93.4
			\tabularnewline
			\cite{Li18}+TASN~\cite{Zheng19}+{\bf VTNs}&ResNet-50 &\underline{\bf{91.2}} &\underline{\bf{95.9}} &\underline{\bf{94.5}}\tabularnewline			 
			\hlinewd{0.8pt}
		\end{tabular}
	\end{center}
	\vspace{-5pt}
	\caption{Accuracy of VTNs compared to the state-of-the-art methods on fine-grained image recognition benchmarks (CUB-Birds~\cite{WelinderEtal2010}, Stanford-Cars~\cite{Krause15}, and FGVC-Aircraft~\cite{Maji13}).}\label{tab:3}\vspace{-20pt}
\end{table}
\begin{table}[t]
	\begin{center}
		\begin{tabular}{ >{\raggedright}m{0.2\linewidth}
				>{\centering}m{0.16\linewidth}
				>{\centering}m{0.16\linewidth} 
				>{\centering}m{0.16\linewidth}
				>{\centering}m{0.16\linewidth}}
			\hlinewd{0.8pt}
			Super Class &ResNet~\cite{He16} &SSN~\cite{Recasens18} &TASN~\cite{Zheng19} &\cite{Li18}+{\bf VTNs} \tabularnewline
			\hline
			\hline
			Plantae &60.3 &63.9 &66.6 &\underline{\bf{68.6}} \tabularnewline
			Insecta &69.1 &74.7 &77.6 &\underline{\bf{79.1}} \tabularnewline
			Aves &59.1 &68.2 &72.0 &\underline{\bf{72.9}} \tabularnewline
			Reptilia &37.4 &43.9 &46.4 &\underline{\bf{48.1}} \tabularnewline
			Mammalia &50.2 &55.3 &57.7 &\underline{\bf{60.6}} \tabularnewline
			Fungi &62.5 &64.2 &70.3 &\underline{\bf{72.1}} \tabularnewline
			Amphibia &41.8 &50.2 &51.6 &\underline{\bf{53.9}} \tabularnewline
			Mollusca &56.9 &61.5 &64.7 &\underline{\bf{66.3}} \tabularnewline
			Animalia &64.8 &67.8 &71.0 &\underline{\bf{73.2}} \tabularnewline
			Arachnida &64.8 &73.8 &75.1 &\underline{\bf{78.2}} \tabularnewline
			Actinoopterygii &57.0 &60.3 &65.5 &\underline{\bf{68.4}} \tabularnewline
			Chromista &57.6 &57.6 &62.5 &\underline{\bf{64.0}} \tabularnewline
			Protozoa &78.1 &79.5 &79.5 &\underline{\bf{81.1}} \tabularnewline
			\hline
			Total &58.4 &63.1 &66.2 &\underline{\bf{68.2}} \tabularnewline
			\hlinewd{0.8pt}
		\end{tabular}
	\end{center}
	\vspace{-5pt}
	\caption{Accuracy of VTNs compared to the state-of-the-art methods on iNaturalist-2017~\cite{Horn18}.}\label{tab:4}\vspace{-20pt}
\end{table}
\begin{table}[t]
	\begin{center}
		\begin{tabular}{ >{\raggedright}m{0.44\linewidth}
				>{\centering}m{0.14\linewidth}
				>{\centering}m{0.14\linewidth}}
			\hlinewd{0.8pt}
			Methods &top-1 err. &top-5 err. \tabularnewline
			\hline
			\hline
			GoogLeNet+GAP~\cite{Zhou16} &59.00 &-\tabularnewline
			VGGNet+ACoL~\cite{Zhang18} &54.08 &43.49\tabularnewline
			ResNet+GCAM~\cite{Selvaraju17} &53.42 &43.12\tabularnewline
			ResNet+STNs~\cite{Jaderberg15}+GCAM~\cite{Selvaraju17} &54.21 &43.33\tabularnewline
			\hline
			ResNet+{\bf VTNs}+GCAM~\cite{Selvaraju17} &\underline{\bf{52.18}} &\underline{\bf{41.76}}\tabularnewline
			\hlinewd{0.8pt}
		\end{tabular}
	\end{center}
	\vspace{-5pt}
	\caption{Localization errors on CUB-Birds~\cite{WelinderEtal2010}.}\label{tab:5}\vspace{-20pt}
\end{table}
\begin{figure}[t]
	\centering
	\renewcommand{\thesubfigure}{}
	\subfigure[(a)]
	{\includegraphics[width=0.122\linewidth]{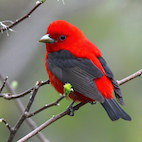}}\hfill
	\subfigure[(b)]
	{\includegraphics[width=0.122\linewidth]{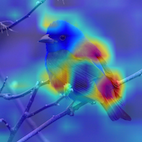}}\hfill
	\subfigure[(c)]
	{\includegraphics[width=0.122\linewidth]{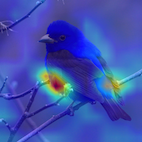}}\hfill
	\subfigure[(d)]
	{\includegraphics[width=0.122\linewidth]{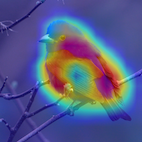}}\hfill
	\subfigure[(e)]
	{\includegraphics[width=0.122\linewidth]{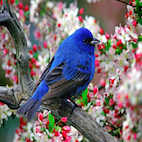}}\hfill
	\subfigure[(f)]
	{\includegraphics[width=0.122\linewidth]{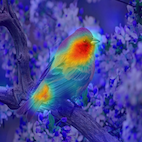}}\hfill
	\subfigure[(g)]
	{\includegraphics[width=0.122\linewidth]{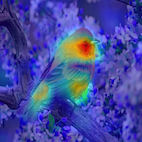}}\hfill
	\subfigure[(h)]
	{\includegraphics[width=0.122\linewidth]{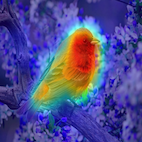}}\hfill\\
	\vspace{-10pt}
	\caption{Network visualization using Grad-CAM~\cite{Selvaraju17}: (a), (e) input images, (b), (f) ResNet-50~\cite{He16}, (c), (g) ResNet-50 with STNs~\cite{Jaderberg15}, and (d), (h) ResNet-50 with VTNs.}\label{img:10}\vspace{-10pt}
\end{figure}
\begin{figure*}[t!]
	\centering
	\renewcommand{\thesubfigure}{}
	\subfigure[]
	{\includegraphics[width=0.122\linewidth]{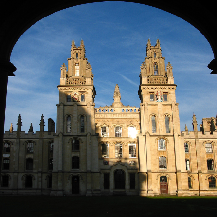}}\hfill
	\subfigure[]
	{\includegraphics[width=0.122\linewidth]{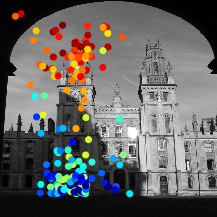}}\hfill
	\subfigure[]
	{\includegraphics[width=0.122\linewidth]{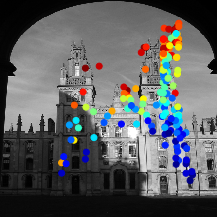}}\hfill
	\subfigure[]
	{\includegraphics[width=0.122\linewidth]{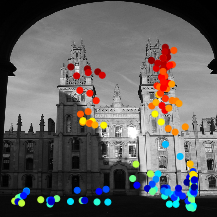}}\hfill
	\subfigure[]
	{\includegraphics[width=0.122\linewidth]{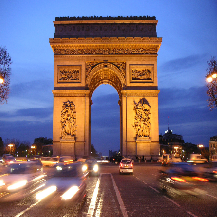}}\hfill
	\subfigure[]
	{\includegraphics[width=0.122\linewidth]{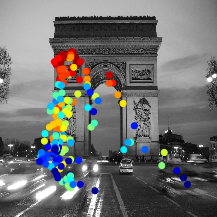}}\hfill
	\subfigure[]
	{\includegraphics[width=0.122\linewidth]{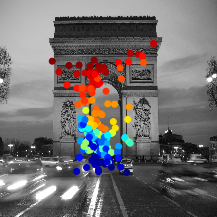}}\hfill
	\subfigure[]
	{\includegraphics[width=0.122\linewidth]{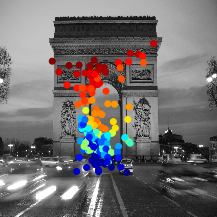}}\hfill\\
	\vspace{-20.5pt}
	\subfigure[]
	{\includegraphics[width=0.122\linewidth]{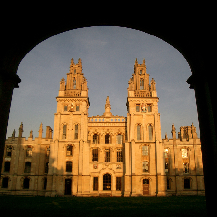}}\hfill
	\subfigure[]
	{\includegraphics[width=0.122\linewidth]{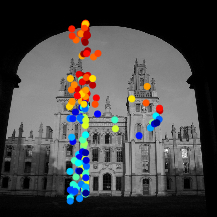}}\hfill
	\subfigure[]
	{\includegraphics[width=0.122\linewidth]{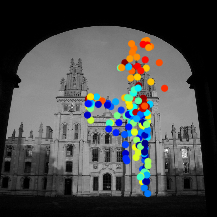}}\hfill
	\subfigure[]
	{\includegraphics[width=0.122\linewidth]{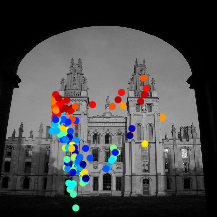}}\hfill
	\subfigure[]
	{\includegraphics[width=0.122\linewidth]{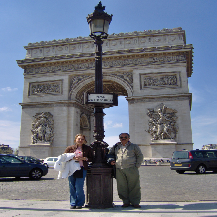}}\hfill
	\subfigure[]
	{\includegraphics[width=0.122\linewidth]{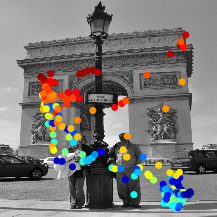}}\hfill
	\subfigure[]
	{\includegraphics[width=0.122\linewidth]{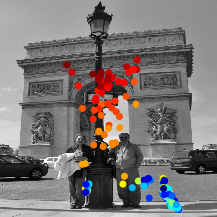}}\hfill
	\subfigure[]
	{\includegraphics[width=0.122\linewidth]{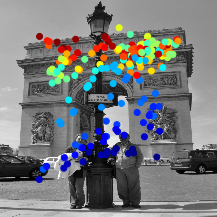}}\hfill\\
	\vspace{-10pt}
	\caption{Visualization of warping fields of VTNs at each channel on some instances of the $\mathcal{R}$Oxford and $\mathcal{R}$Paris benchmarks~\cite{Radenovic18}. VTNs not only localize different semantic parts in different channels but
		identify the same points across different images.}\label{img:11}\vspace{-10pt}
\end{figure*}
\begin{table}[t]
	\begin{center}
		\begin{tabular}{ >{\raggedright}m{0.36\linewidth}
				>{\centering}m{0.1\linewidth}
				>{\centering}m{0.1\linewidth} 
				>{\centering}m{0.1\linewidth}
				>{\centering}m{0.1\linewidth}}
			\hlinewd{0.8pt}
			\multirow{2}{*}{Methods}
			&\multicolumn{2}{ c }{Medium} &\multicolumn{2}{ c }{Hard} \tabularnewline
			\cline{2-5}
			&$\mathcal{R}$Oxf &$\mathcal{R}$Par &$\mathcal{R}$Oxf &$\mathcal{R}$Par \tabularnewline
			\hline
			\hline
			Pretr.+MAC~\cite{Tolias16} &41.7 &66.2 &18.0 &44.1 \tabularnewline
			Pretr.+SPoC~\cite{Babenko15} &39.8 &69.2 &12.4 &44.7 \tabularnewline
			Pretr.+CroW~\cite{Kalantidis16} &42.4 &70.4 &13.3 &47.2 \tabularnewline
			Pretr.+GeM~\cite{Radenovi19tpami} &45.0 &70.7 &17.7 &48.7 \tabularnewline
			Pretr.+R-MAC~\cite{Tolias16} &49.8 &74.0 &18.5 &52.1 \tabularnewline
			DELF~\cite{Noh17,Tolias16,Philbin07} &67.8 &76.9 &43.1 &55.4 \tabularnewline
			\cite{Radenovi19tpami}+GeM &64.7 &77.2 &38.5 &56.3 \tabularnewline
			\cite{Gordo17}+R-MAC &60.9 &78.9 &32.4 &59.4 \tabularnewline
			\cite{Gordo17}+R-MAC+STNs~\cite{Jaderberg15} &61.3 &79.4 &36.1 &59.8 \tabularnewline
			\hline
			DELF~\cite{Noh17,Tolias16,Philbin07}+{\bf VTNs} &\underline{\bf{69.7}} &78.1 &45.1 &56.4 \tabularnewline
			\cite{Radenovi19tpami}+GeM+{\bf VTNs} &67.4 &80.5 &\underline{\bf{45.5}} &57.1 \tabularnewline
			\cite{Gordo17}+R-MAC+{\bf VTNs} &65.6 &\underline{\bf{82.7}} &43.3 &\underline{\bf{60.9}} \tabularnewline			
			\hlinewd{0.8pt}
		\end{tabular}
	\end{center}
	\vspace{-5pt}
	\caption{Comparison of VTNs with the state-of-the-art methods on the $\mathcal{R}$Oxford and $\mathcal{R}$Paris benchmarks~\cite{Radenovic18}.}\label{tab:6}\vspace{-20pt}
\end{table}

\noindent{{\bf Incorporating second-order pooling strategies.}} 
While our VTNs can be used on their own, they can also integrate second-order pooling schemes, such as bilinear pooling (BP)~\cite{Lin15}, compact BP (CBP)~\cite{Gao16}, kernel pooling (KP)~\cite{Cui17}, matrix power normalized covariance pooling (MPN-COV)~\cite{Li17}, and iterative matrix square root normalization of covariance pooling (iSQRT-COV)~\cite{Li18}, which yield state-of-the-art results on fine-grained image recognition. In this set of experiments, we use ResNet-50~\cite{He16} as backbone. 
%We applied VTNs to a last convolution layer before the pooling layers.
As shown in Table~\ref{tab:2}, our VTNs consistently outperform the corresponding pooling strategy on its own, thus confirming the benefits of using channel-wise warped regions.
%\MS{Or is this due to the fact that we have more parameters?} 
\vspace{+10pt}

\noindent{{\bf Comparison with the state-of-the-art methods.}} 
We also compare VTNs with the state-of-the-art fine-grained image recognition methods, such as RA-CNN~\cite{Fu17}, MA-CNN~\cite{Zheng17}, DFL-CNN~\cite{Wang18}, DT-RAM~\cite{Li17}, MAMC~\cite{Sun18}, NTSN~\cite{Sun18}, DCL~\cite{Chen19}, and TASN~\cite{Zheng19}.
Since our VTN is designed as a generic drop-in layer that can be combined with existing backbones and pooling strategies, we report the results of VTNs combined with DCL~\cite{Chen19}, TASN~\cite{Zheng19}, and iSQRT-COV~\cite{Li18}, which are the top-performers on this task. 
%Note that we designed VTNs as a generic drop-in layer, which can be combined with most methods that use a standard backbone network such as VGGNet~\cite{Simonyan15} and ResNet~\cite{He16}, and thus it is hard to directly compare them with other methods. \MS{Explain why. What kind of backbone do they use?} We instead used VTNs with iSQRT-COV~\cite{Li18}, which was the top in the ablation study. 
As can be seen in Table~\ref{tab:3}, our method outperforms the state of the art in most cases. In Table~\ref{tab:4}, we further evaluate VTNs with iSQRT-COV~\cite{Li18} on iNaturalist-2017~\cite{Horn18}, the largest fine-grained recognition dataset, on which we consistently outperform the state of the art. \vspace{+10pt}

\noindent{{\bf Network visualization.}} 
To analyze the feature representation capability of VTNs, we applied Grad-CAM (GCAM)~\cite{Selvaraju17}, which uses gradients to calculate the importance of the spatial locations, to STN- and VTN-based networks. As shown in \figref{img:10} and Table~\ref{tab:5}, compared to the ResNet-50~\cite{He16} backbone, STNs~\cite{Jaderberg15} only focus on the most discriminative parts, and thus discard other important parts. Unlike them, VTNs improve the feature representation power by allowing the networks to focus on the most discriminative parts represented by each feature channel. \vspace{-5pt} 

\subsection{Instance-level Image Retrieval}
Finally, we evaluate our VTNs on the task of instance-level image retrieval using the $\mathcal{R}$Oxford and $\mathcal{R}$Paris benchmarks~\cite{Radenovic18}, which address some limitations of the standard Oxford-5K~\cite{Philbin07} and Paris-6K benchmarks~\cite{Philbin08}, such as annotation errors, size of the dataset, and level of difficulty, and comprise 4,993 and 6,322 images, respectively. Following standard practice, we use the mean average precision (mAP)~\cite{Philbin07} for quantitative evaluation. We follow the evaluation protocol of~\cite{Radenovic18}, using two evaluation setups (\emph{Medium} and \emph{Hard}). As baselines, we use a pretrained ResNet-50~\cite{He16} backbone, followed by various pooling methods, such as MAC~\cite{Tolias16}, SPoC~\cite{Babenko15}, CroW~\cite{Kalantidis16}, R-MAC~\cite{Tolias16}, and GeM~\cite{Radenovi19tpami}. We also evaluate deep local attentive features (DELF)~\cite{Noh17} with an aggregated selective match kernel~\cite{Tolias16} and spatial verification~\cite{Philbin07} that learns spatial attention, and incorporate our VTNs into them. Furthermore, we report the results of end-to-end training techniques~\cite{Radenovi19tpami,Gordo17}, and incorporate our VTNs on top of them. As evidenced by our significantly better results in Table~\ref{tab:6}, focusing on the most discriminative parts at each feature channel is one of the key to the success of instance-level image retrieval. Note that the comparison with STNs shows the benefits of our approach, which accounts for different semantic concepts across different feature channels and thus, even for rigid objects, is able to learn more discriminative feature representations than a global warping. \figref{img:11} visualizes some warping fields of VTNs. \vspace{-5pt}

\section{Conclusion}
We have introduced VTNs that predict channel-wise warping fields to boost the representation power of an intermediate CNN feature map by reconfiguring the features spatially and channel-wisely. VTNs account for the fact that the individual feature channels can represent different semantic information and require different spatial transformations. To this end, we have developed an encoder-decoder network that relies on channel squeeze and expansion modules to account for inter-channel relationships. To improve the localization ability of the predicted warping fields, we have further introduced a loss function defined between the warped features of pairs of instances. Our experiments have shown that VTNs consistently boost the features' representation power and consequently the networks' accuracy on fine-grained image recognition and instance-level image retrieval tasks. In the future, we will aim to apply VTNs to other tasks, such as person re-identification and local feature matching. 

\section*{Acknowledgments}
This work was supported in part by the Swiss National Science Foundation via the Sinergia grant CRSII5-180359. The work of S. Kim was supported by Institute for Information \& communications Technology Planning \& Evaluation (IITP) grant funded by the Korea government (MSIT) (No. 2020-0-00368, A Neural-Symbolic Model for Knowledge Acquisition and Inference Techniques).

\clearpage
\bibliographystyle{splncs04}
\bibliography{main}

\begin{thebibliography}{10}
\providecommand{\url}[1]{\texttt{#1}}
\providecommand{\urlprefix}{URL }
\providecommand{\doi}[1]{https://doi.org/#1}

\bibitem{Arandjelovic13}
Arandjelovic, R., Zisserman, A.: All about vlad. In: CVPR  (2013)

\bibitem{Babenko15}
Babenko, A., Lempitsky, V.: Aggregating deep convolutional features for image
  retrieval. In: ICCV  (2015)

\bibitem{Bahdanau15}
Bahdanau, D., Cho, K., Bengio, Y.: Neural machine translation by jointly
  learning to align and translate. In: ICLR  (2015)

\bibitem{Battaglia16}
Battaglia, P.W., Pascanu, R., Lai, M., Rezende, D.J., Kavukcuoglu, K.:
  Interaction networks for learning about objects, relations and physics. In:
  NeurIPS  (2016)

\bibitem{Bau17}
Bau, D., Zhou, B., Khosla, A., Oliva, A., Torralba, A.: Network dissection:
  Quantifying interpretability of deep visual representations. In: CVPR  (2017)

\bibitem{Berg14}
Berg, T., Liu, J., Lee, S.W., Alexander, M.L., Jacobs, D.W., Belhumeur, P.N.:
  Birdsnap: Large-scale fine-grained visual categorization of birds. In: CVPR
  (2014)

\bibitem{Chen17}
Chen, L., Zhang, H., Xiao, J., Nie, L., Shao, J., Liu, W., Chua, T.: Sca-cnn:
  Spatial and channel-wise attention in convolutional networks for image
  captioning. In: CVPR  (2017)

\bibitem{Chen19}
Chen, Y., Bai, Y., Zhang, W., Mei, T.: Destruction and construction learning
  for fine-grained image recognition. In: CVPR  (2019)

\bibitem{Cui17}
Cui, Y., Zhou, F., Wang, J., Liu, X., Lin, Y., Belongie, S.: Kernel pooling for
  convolutional neural networks. In: CVPR  (2017)

\bibitem{Dai17}
Dai, J., Qi, H., Xiong, Y., Li, Y., Zhang, G., Hu, H., Wei, Y.: Deformable
  convolutional networks. In: ICCV  (2017)

\bibitem{imagenet_cvpr09}
Deng, J., Dong, W., Socher, R., Li, L.J., Li, K., Fei-Fei, L.: {ImageNet: A
  Large-Scale Hierarchical Image Database}. In: CVPR  (2009)

\bibitem{Esteves18}
Esteves, C., Allen-Blanchette, C., Zhou, X., Daniilidis, K.: Polar transformer
  networks. In: ICLR  (2018)

\bibitem{Everingham15}
Everingham, M., Eslami, S.M.A., Van~Gool, L., Williams, C.K.I., Winn, J.,
  Zisserman, A.: The pascal visual object classes challenge: A retrospective.
  IJCV  \textbf{111}(1),  98 -- 136 (2015)

\bibitem{Felzenszwalb10}
Felzenszwalb, P.F., Girshick, R.B., McAllester, D., Ramanan, D.: Object
  detection with discriminative trained part based models. IEEE Trans. PAMI
  \textbf{32}(9),  1627 -- 1645 (2010)

\bibitem{Fu19}
Fu, J., Liu, J., Tian, H., Fang, Z., Lu, H.: Dual attention network for scene
  segmentation. In: CVPR  (2019)

\bibitem{Fu17}
Fu, J., Zheng, H., Mei, T.: Look closer to see better: Recurrent attention
  convolutional neural network for fine-grained image recognition. In: CVPR
  (2017)

\bibitem{Gao16}
Gao, Y., Beijbom, O., Zhang, N., Darrell, T.: Compact bilinear pooling. In:
  CVPR  (2016)

\bibitem{Garcia18}
Gonzalez-Garcia, A., Modolo, D., Ferrari, V.: Do semantic parts emerge in
  convolutional neural networks? IJCV  \textbf{126}(5),  476--494 (2018)

\bibitem{Gordo17}
Gordo, A., Almaz\`{a}n, J., Revaud, J., Larlus, D.: End-to-end learning of deep
  visual representations for image retrieval. IJCV  \textbf{124}(2),  237 --
  254 (2017)

\bibitem{Gregor15}
Gregor, K., Danihelka, I., Graves, A., Rezende, D.J., Wierstra, D.: Draw: A
  recurrent neural network for image generation. In: ICML  (2015)

\bibitem{He16}
He, K., Zhang, X., Ren, S., Sun, J.: Deep residual learning for image
  recognition. In: CVPR  (2016)

\bibitem{Horn18}
Horn, G.V., Aodha, O.M., Song, Y., Cui, Y., Sun, C., Shepard, A., Adam, H.,
  Perona, P., Belongie, S.: The inaturalist species classification and
  detection dataset. In: CVPR  (2018)

\bibitem{Hu18}
Hu, J., Shen, L., Sun, G.: Squeeze-and-excitation networks. In: CVPR  (2018)

\bibitem{huang2017densely}
Huang, G., Liu, Z., Laurens, V.D.M., Weinberger, K.Q.: Densely connected
  convolutional networks. In: CVPR  (2017)

\bibitem{Huang16}
Huang, S., Xu, Z., Tao, D., Zhang, Y.: Part-stacked cnn for fine-grained visual
  categorization. In: CVPR  (2016)

\bibitem{Ioffe15}
Ioffe, S., Szegedy, C.: Batch normalization: Accelerating deep network training
  by reducing internal covariate shift. In: ICML  (2015)

\bibitem{Itti98}
Itti, L., Koch, C., Niebur, E.: A model of saliency-based visual attention for
  rapid scene analysis. TPAMI  \textbf{20}(11),  1254 -- 1259 (1998)

\bibitem{Jaderberg15}
Jaderberg, M., Simonyan, K., Zisserman, A., Kavukcuoglu, K.: Spatial
  transformer networks. In: NeurIPS  (2015)

\bibitem{Jeong19}
Jeong, J., Shin, J.: Training cnns with selective allocation of channels. In:
  ICML  (2019)

\bibitem{Kalantidis16}
Kalantidis, Y., Mellina, C., Osindero, S.: Cross-dimensional weighting for
  aggregated deep convolutional features. In: ECCVW  (2016)

\bibitem{Kim19}
Kim, S., Min, D., Jeong, S., Kim, S., Jeon, S., Sohn, K.: Semantic attribute
  matching networks. In: CVPR  (2019)

\bibitem{Krause15}
Krause, J., Stark, M., Deng, J., Fei-Fei, L.: 3d object representations for
  fine-grained categorization. In: CVPR  (2015)

\bibitem{Krizhevsky12}
Krizhevsky, A., Sutskever, I., Hinton, G.E.: Imagenet classification with deep
  convolutional neural networks. In: NeurIPS  (2012)

\bibitem{Larochelle10}
Larochelle, H., Hinton, G.E.: Learning to combine foveal glimpses with a
  third-order boltzmann machine. In: NeurIPS  (2010)

\bibitem{Li18}
Li, P., Xie, J., Wang, Q., Zuo, W.: Towards faster training of global
  covariance pooling networks by iterative matrix square root normalization.
  In: CVPR  (2018)

\bibitem{Li17}
Li, Z., Yang, Y., Liu, X., Zhou, F., Wen, S., Xu, W.: Dynamic computational
  time for visual attention. In: ICCVW  (2017)

\bibitem{Lin17inv}
Lin, C.H., Lucey, S.: Inverse compositional spatial transformer networks. In:
  CVPR  (2017)

\bibitem{Lin15}
Lin, T.Y., RoyChowdhury, A., Maji, S.: Bilinear cnn models for fine-grained
  visual recognition. In: ICCV  (2015)

\bibitem{Lowe04}
Lowe, D.G.: Distinctive image features from scale-invariant keypoints. IJCV
  \textbf{60}(2),  91--110 (2004)

\bibitem{Maji13}
Maji, S., Kannala, J., Rahtu, E., Blaschko, M., Vedaldi, A.: Fine-grained
  visual classification of aircraft. Tech. rep. (2013)

\bibitem{Mikolajczyk05}
Mikolajczyk, K., Schmid, C.: A performance evaluation of local descriptors.
  TPAMI  \textbf{27}(10),  1615 -- 1630 (2005)

\bibitem{Noh17}
Noh, H., Araujo, A., Sim, J., Weyand, T., Han, B.: Large-scale image retrieval
  with attentive deep local features. In: ICCV  (2017)

\bibitem{paszke2017automatic}
Paszke, A., Gross, S., Chintala, S., Chanan, G., Yang, E., DeVito, Z., Lin, Z.,
  Desmaison, A., Antiga, L., Lerer, A.: Automatic differentiation in pytorch
  (2017)

\bibitem{Philbin07}
Philbin, J., Chum, O., Isard, M., Sivic, J., Zisserman, A.: Object retrieval
  with large vocabularies and fast spatial matching. In: CVPR  (2007)

\bibitem{Philbin08}
Philbin, J., Chum, O., Isard, M., Sivic, J., Zisserman, A.: Lost in
  quantization: Improving particular object retrieval in large scale image
  databases. In: CVPR  (2008)

\bibitem{Radenovic18}
Radenovi\`{c}, F., Iscen, A., Tolias, G., Avrithis, Y., Chum, O.: Revisiting
  oxford and paris: Large-scale image retrieval benchmarking. In: CVPR  (2018)

\bibitem{Radenovi19tpami}
Radenovi\`{c}, F., Tolias, G., Chum, O.: Fine-tuning cnn image retrieval with
  no human annotation. TPAMI  \textbf{41}(7),  1655--1668 (2019)

\bibitem{Recasens18}
Recasens, A., Kellnhofer, P., Stent, S., Matusik, W., Torralba, A.: Learning to
  zoom: a saliency-based sampling layer for neural networks. In: ECCV  (2018)

\bibitem{Ren15}
Ren, S., He, K., Girshick, R., Sun, J.: Faster r-cnn: Towards real-time object
  detection with region proposal networks. In: NeurIPS  (2015)

\bibitem{Rocco18}
Rocco, I., Arandjelovi\'c, R., Sivic, J.: End-to-end weakly-supervised semantic
  alignment. In: CVPR  (2018)

\bibitem{Rodriguez18}
Rodriguez, P., Gonfaus, J.M., Cucurull, G., Xavierroca, F., Gonzalez, J.:
  Attend and rectify: A gated attention mechanism for fine-grained recovery.
  In: ECCV  (2018)

\bibitem{Santoro17}
Santoro, A., Raposo, D., Barrett, D.G., Malinowski, M., Pascanu, R., Battaglia,
  P., Lillicrap, T.: A simple neural network moduel for relational reasoning.
  In: NeurIPS  (2017)

\bibitem{Selvaraju17}
Selvaraju, R.R., Cogswell, M., Das, A., Vedantam, R.: Grad-cam: Visual
  explanations from deep networks via gradient-based localization. In: ICCV
  (2017)

\bibitem{Simo-Serra15b}
Simo-Serra, E., Trulls, E., Ferraz, L., Kokkinos, I., Fua, P., Moreno-Noguer,
  F.: Discriminative learning of deep convolutional feature point descriptors.
  In: ICCV  (2015)

\bibitem{Simonyan15}
Simonyan, K., Zisserman, A.: Very deep convolutional networks for large-scale
  image recognition. In: ICLR  (2015)

\bibitem{Sivic03}
Sivic, J., Zisserman, A.: Video google: A text retrieval approach to object
  matching in videos. In: ICCV  (2003)

\bibitem{Sun19}
Sun, K., Xiao, B., Liu, D.: Deep high-resolution representation learning for
  human pose estimation. In: CVPR  (2019)

\bibitem{Sun18}
Sun, M., Yuan, Y., Zhou, F., Ding, E.: Multi-attention multi-class constraint
  for fine-grained image recognition. In: ECCV  (2018)

\bibitem{Tolias16}
Tolias, G., Avrithis, Y., Jégou, H.: Image search with selective match
  kernels: Aggregation across single and multiple images. IJCV
  \textbf{116}(3),  247--261 (2016)

\bibitem{Tompson15}
Tompson, J., Goroshin, R., Jain, A., LeCun, Y., Bregler, C.: Efficient object
  localization using convolutional networks. In: CVPR  (2015)

\bibitem{Vaswani17}
Vaswani, A., Shazeer, N., Parmar, N., Uszkoreit, J., Jones, L., Gomez, A.N.,
  Kaiser, L., Polosukhin, I.: Attention is all you need. In: NeurIPS  (2017)

\bibitem{Wang17}
Wang, F., Jiang, M., Qian, C., Yang, S., Li, C., Zhang, H., X., W.: Residual
  attention network for image classification. In: CVPR  (2017)

\bibitem{Wang18}
Wang, Y., Morariu, V.I., Davis, L.S.: Learning a discriminative filter bank
  within a cnn for fine-grained recognition. In: CVPR  (2018)

\bibitem{WelinderEtal2010}
Welinder, P., Branson, S., Mita, T., Wah, C., Schroff, F., Belongie, S.,
  Perona, P.: Caltech-ucsd birds 200. Tech. Rep. CNS-TR-2010-001, California
  Institute of Technology (2010)

\bibitem{Woo18}
Woo, S., Park, J., Lee, J.Y., Kweon, I.S.: Cbam: Convolutional block attention
  module. In: ECCV  (2018)

\bibitem{Wu18}
Wu, Y., He, K.: Group normalization. In: ECCV  (2018)

\bibitem{Xu15}
Xu, K., Ba, J., Kiros, R., Cho, K., Courville, A., Salakhudinov, R., Zemel, R.,
  Bengio, Y.: Show, attend and tell: Neural image caption generation with
  visual attention. In: ICML  (2015)

\bibitem{Xu18}
Xu, T., Zhang, P., Huang, Q., Zhang, H., Gan, Z., Huang, X., He, X.: Attngan:
  Fine-grained text to image generation with attentional generative adversarial
  networks. In: CVPR  (2018)

\bibitem{Yi16}
Yi, K.M., Trulls, E., Lepetit, V., Fua, P.: Lift: Learned invariant feature
  transform. In: ECCV  (2016)

\bibitem{Zhang18}
Zhang, H., Dana, K., Shi, J., Zhang, Z., Wang, X., Tyagi, A., Agrawal, A.:
  Context encoding for semantic segmentation. In: CVPR  (2018)

\bibitem{Zhang19}
Zhang, H., Goodfellow, I., Metaxas, D., Odena, A.: Self-attention generative
  adversarial networks. In: ICML  (2019)

\bibitem{Zheng17}
Zheng, H., Fu, J., Mei, T., Luo, J.: Learning multi-attention convolutional
  neural network for fine-grained image recognition. In: ICCV  (2017)

\bibitem{Zheng19}
Zheng, H., Fu, J., Zha, Z.J., Luo, J.: Looking for the devil in the details:
  Learning trilinear attention sampling network for fine-grained image
  recognition. In: CVPR  (2019)

\bibitem{Zheng15}
Zheng, L., Shen, L., Tian, L., Wang, S., Wang, J., Tian, Q.: Scalable person
  re-identification: A benchmark. In: ICCV  (2015)

\bibitem{Zhou16}
Zhou, B., Khosla, A., Lapedriza, A., Oliva, A., Torralba, A.: Learning deep
  features for discriminative localization. In: CVPR  (2016)

\bibitem{Zhu19}
Zhu, X., Cheng, D., Zhang, Z., Lin, S., Dai, J.: An empirical study of spatial
  attention mechanisms in deep networks. In: ICCV  (2019)

\end{thebibliography}
\end{document}